\documentclass[letterpaper]{article}

\usepackage{natbib,alifeconf}  
\usepackage{url,hyperref,cleveref}
\usepackage{booktabs}
\usepackage{tcolorbox}

\usepackage{tcolorbox}
\tcbuselibrary{skins, breakable}

\tcbset{
  enhanced,
  breakable,                
  width=\columnwidth,       
  before skip=1em,          
  after skip=1em,           
  colback=gray!5,
  colframe=gray!50,
  fonttitle=\bfseries,
  coltitle=black,
  center title,
  rounded corners,
  fontupper=\ttfamily,
}
\title{A Concurrent Modular Agent: Framework for Autonomous LLM Agents}
\author{
    Norihiro Maruyama$^{1,2}$,
    Takahide Yoshida$^{2}$, 
    Hiroki Sato$^{2}$,
    Atsushi Masumori$^{1,2}$,
    Johnsmith$^{2}$\and
    Takashi Ikegami$^{1,2}$ \\
    \mbox{}\\
    $^1$Alternative Machine Inc.\\
    $^2$The University of Tokyo, Komaba, Japan \\
    maruyama@sacral.c.u-tokyo.ac.jp
} 

\begin{document}

\maketitle
\begin{abstract}
We introduce the \textit{Concurrent Modular Agent} (CMA), a framework that orchestrates multiple Large-Language-Model (LLM)-based modules that operate fully asynchronously yet maintain a coherent and fault-tolerant behavioral loop. This framework addresses long-standing difficulties in agent architectures by letting intention emerge from language-mediated interactions among autonomous processes. This approach enables flexible, adaptive, and context-dependent behavior through the combination of concurrently executed modules that offload reasoning to an LLM, inter-module communication, and a single shared global state.We consider this approach to be a practical realization of Minsky's Society of Mind theory. We demonstrate the viability of our system through two practical use-case studies. The emergent properties observed in our system suggest that complex cognitive phenomena like self-awareness may indeed arise from the organized interaction of simpler processes, supporting Minsky-Society of Mind concept and opening new avenues for artificial intelligence research. The source code for our work is available at: \url{https://github.com/AlternativeMachine/concurrent-modular-agent}.
\end{abstract}

\section{Introduction}
Natural biological systems exhibit inherently asynchronous and embodied intelligence\citep{pfeifer2006body,brooks1991intelligence,ackley2013}. Animals do not rely on centralized, symbolic reasoning alone, but instead coordinate multiple loosely coupled processes across time and space\citep{som}. For instance, they can monitor the environment for threats while eating, explore unfamiliar surroundings, and plan future actions—all concurrently. In contrast, most current AI systems, including large language model (LLM)-based agents, remain largely synchronous and disembodied. Addressing this gap requires rethinking system architectures from the ground up.

Concurrent processing is essential when constructing autonomous intelligent systems. Traditional AI architectures have long grappled with this issue\citep{pfeifer1995}, leading to numerous proposals such as Brooks’ subsumption architecture\citep{brooks1986}, which conceptualizes intelligent behavior as a hierarchy of simple, reactive modules. Similarly, blackboard systems organize shared information spaces where multiple agents cooperate. Modern robotic middleware, such as the Robot Operating System (ROS)\citep{ROS}, supports asynchronous execution through message passing between distributed nodes.


Large Language Models offer new affordances \citep{bubeck2023sparks}. Their use of natural language for both input and output enables general-purpose communication among functional components, and recent work has explored multi-agent reasoning by orchestrating LLMs through structured prompts. This capability has been explored in recent studies such as Project-Sid and Lyfe Agent\citep{sid2024, lyfe2024}. In Project-Sid, each LLM concurrent module is stateless and writes its output to a common database; inter-module information is transported through the database via retrieval and storage. Lyfe Agent particularly emphasises memory structures, incorporating both short-term and long-term memory.

Building on these insights, we propose a novel asynchronous and distributed architecture composed of multiple interacting LLM modules. Each module executes independently, communicates using natural language as a common protocol, and maintains its own task logic through prompt engineering. Modules interact concurrently via MQTT-based messaging and share long-term information through a common ChromaDB memory. This architecture allows for flexible, adaptive system behavior and reduces reliance on local computation by offloading reasoning to cloud-based LLM APIs. We consider this approach a practical realization of Minsky’s Society of Mind theory.

To demonstrate the visibility of our system, we apply it to two physical robotic platforms: a mobile plant robot (plantbot) and an android.
While prior android systems have used LLMs in limited ways—primarily by issuing predefined prompts for dialogue or gesture imitation—our implementation distributes control across more than twenty asynchronously interacting modules. This enables richer and more context-sensitive interactions grounded in the android's physical embodiment.
Meanwhile, plantbot is a hybrid lifeform that links a living plant and a mobile robot through LLM-based modules.

The primary contributions of this framework include:
\begin{itemize}
    \item Robustness through modular concurrent processing.
    \item Practically unbounded scalability.
    \item A unification of concurrent module interaction with agent-based modeling via shared databases and MQTT communication.
\end{itemize}

This framework supports the construction of highly flexible and sophisticated LLM-based intelligent systems.
The following sections provide a detailed description of the proposed framework, followed by two practical use-case studies that demonstrate its utility.

\section{Related Work}
Our approach builds upon and extends several threads of research in agent architectures, embodied AI, and artificial life. 

Classical robotic architectures such as Brooks' subsumption architecture conceptualized behavior as layered, reactive modules with inhibitory hierarchies \citep{brooks1986}. Blackboard systems \citep{corkill1991} introduced global shared workspaces for cooperative reasoning, while frameworks such as ROS \citep{ROS} provided asynchronous execution across distributed nodes. However, these systems often lacked mechanisms for flexible communication or high-level reasoning.

Recent work in LLM-based agents has explored the use of prompt engineering and memory augmentation to enable multi-step reasoning and planning \citep{wang2024survey,park2023generative}. Projects such as Project-Sid \citep{sid2024} and Lyfe Agent \citep{lyfe2024} have introduced multi-agent LLM societies. SID in particular demonstrated how thousands of agents, coordinated by a central orchestration layer (PIANO), can develop complex behaviors such as role differentiation, cultural transmission, and rule negotiation within a simulated Minecraft environment. These results highlight the scalability and richness of language-mediated agent societies.

In contrast, our system investigates smaller-scale, deeply embodied agents with fully asynchronous modularity. Rather than using LLMs to simulate thousands of minds, we distribute mind-like functions (perception, memory, planning, self-reflection) across LLM modules within a single robotic entity. This mirrors the architectural spirit of Minsky's \textit{Society of Mind} \citep{som}, where intelligence is distributed over specialized, intercommunicating processes.

Our work is also informed by Paulien Hogeweg's perspective on structure-oriented modeling in artificial life \citep{hogeweg1988}, which emphasizes the importance of specifying micro-level interaction rules and observing emergent macro phenomena. Similarly, David Ackley's idea of \textit{indefinitely scalable computation} \citep{ackley2013} inspires our design: instead of global synchrony, our system relies on local, asynchronous interactions among loosely coupled modules to support robust and adaptive behavior.

\section{Proposed  System Architecture}
\begin{figure*}[h]
\centering
\includegraphics[width=\linewidth]{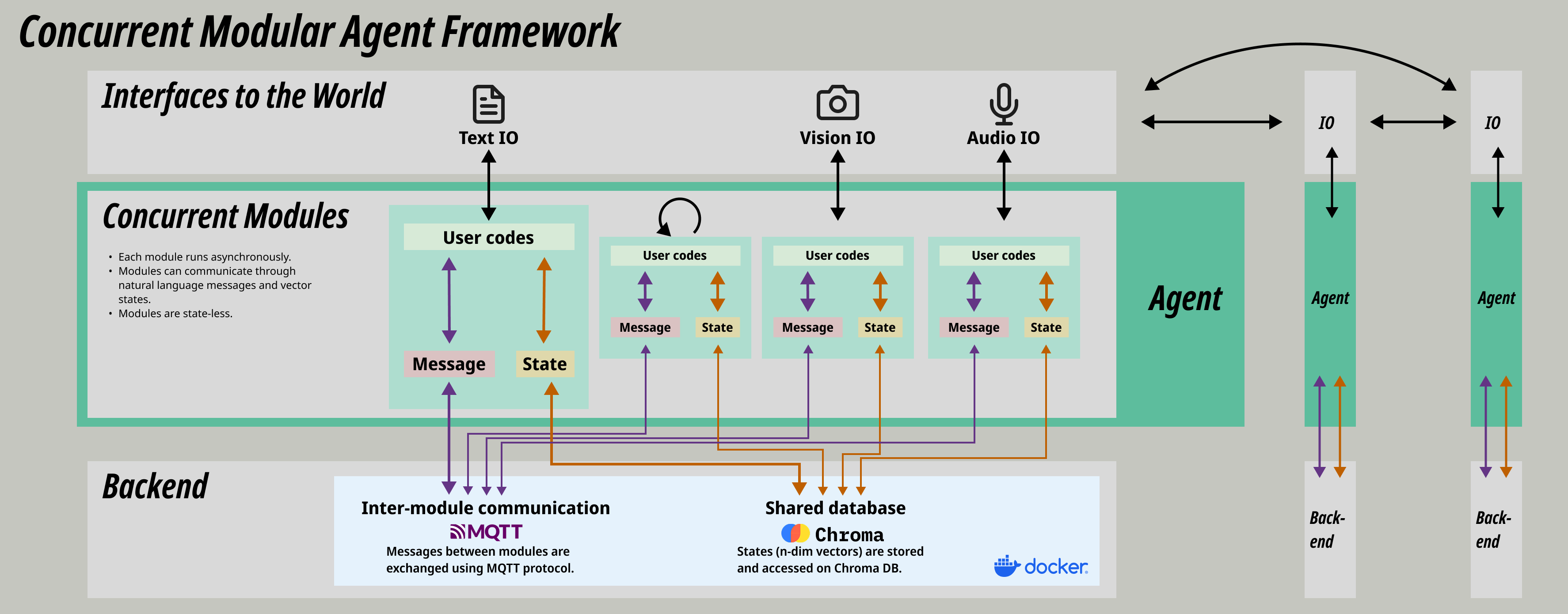}
\caption{Overview of the CMA architecture. Each agent comprises multiple functional modules that run concurrently, exchanging messages and sharing high-dimensional state vectors via shared databases (currently implemented using MQTT and Chroma DB on Docker). Modules may interface with the external world or other agents through dedicated I/O channels (text, vision, audio, etc.), or operate as internal processes.}
\label{framework_overview}
\end{figure*}

We propose a modular agent framework called the \textit{Concurrent Modular Agent} (CMA), in which multiple LLM-based modules operate asynchronously and interact via natural language communication. An overview of the framework is shown in Figure \ref{framework_overview}. As shown in the figure, the system consists of three main components: (1) a set of functional modules, (2) a shared vector-based memory store (global state), and (3) an inter-module communication mechanism.

Each module in the CMA operates independently and asynchronously, allowing the agent to exhibit consistent and robust behaviors akin to those observed in biological organisms or multi-threaded systems. 

Rather than relying on a central control loop or fixed scheduling, each module operates asynchronously and focuses solely on its assigned task. The termination or failure of one module does not directly affect the functioning of the others.
This decentralized execution model allows the agent to, for example, plan, perceive, and act simultaneously—facilitating robust behavior even in dynamic or uncertain environments.
Through the use of shared memory and natural language communication, modules coordinate their perceptions, actions, share observations, forming a loosely coupled yet coherent system.
In the following, we describe each component in detail.

\subsection{Modules}
Each module is implemented as an asynchronous Python function and is designed to perform a specific subtask, such as perception, memory, planning, individual senses, or action execution.
These modules are independently executed and collectively form a coherent agent.
Modules are isolated and do not share internal state but coordinate via a shared memory (global state) and message passing. 
While each module may internally invoke LLMs for reasoning and language generation, the LLM itself is treated as a black-box service accessed via external APIs (e.g., OpenAI’s GPT series, deepseek series).

Each module can perform the following four operations:

\begin{enumerate}
    \item \textbf{Interaction with the external world:} e.g., receiving visual, audio, or text input, or accessing web resources.
    \item \textbf{Retrieval information from global state:} querying the vector store for past knowledge or messages.
    \item \textbf{Storing information into global state:} embedding and saving textual output into shared memory.
    \item \textbf{Sending messages to other modules:} send a text message directly to another module.
\end{enumerate}

Operations (2) and (3) are described in the following subsection; operation (4) is detailed thereafter.

\subsection{Global State}
To support long-term memory of the agent and inter-module knowledge sharing, textual messages are embedded into vector representations and stored in a shared vector database.
All modules are able to asynchronously store information in and retrieve information from the database via queries.
We use ChromaDB, an open-source vector store, to implement fast and flexible retrieval. Modules can query the database to retrieve relevant past utterances or contextual information, enabling reasoning based on both self-generated and externally generated context.
Furthermore, since ChromaDB is run in a docker container and adopts HTTP access mode, each module can, in principle, be executed on a separate host. This enables the scalability of the CMA framework.

Note that this idea—maintaining a single shared state across multiple modules—has been adopted in several previous architectures \citep{sid2024}. However, as will be described in the following section, our framework is novel in that it enables richer and more diverse behaviors by allowing parallel communication among modules.

\subsection{Inter-Module Communication}
Each module is capable of sending text messages to other modules at any time. 
Upon receiving a message, a module can change its behavior depending on the message content --- for example, by initiating or halting execution, triggering specific functions, or modifying its internal processes.
As described earlier, this asynchronous message-passing mechanism for coordination is inspired by the subsumption architecture.

Asynchronous message passing among modules is implemented using MQTT  \citep{MQTT}, a lightweight publish/subscribe protocol. 
Although MQTT is a publish/subscribe-based messaging system, we used it as a backend to implement one-to-one message passing.
As with the global state, the use of MQTT enhances the scalability of the framework. 
MQTT protocol is network-transparent, so it allows modules to be executed independently of their host environment.

\section{Application Example 1: Hybrid Lifeform \textit{Plantbot}}
\begin{figure}[t]
    \centering
    \includegraphics[width=\linewidth]{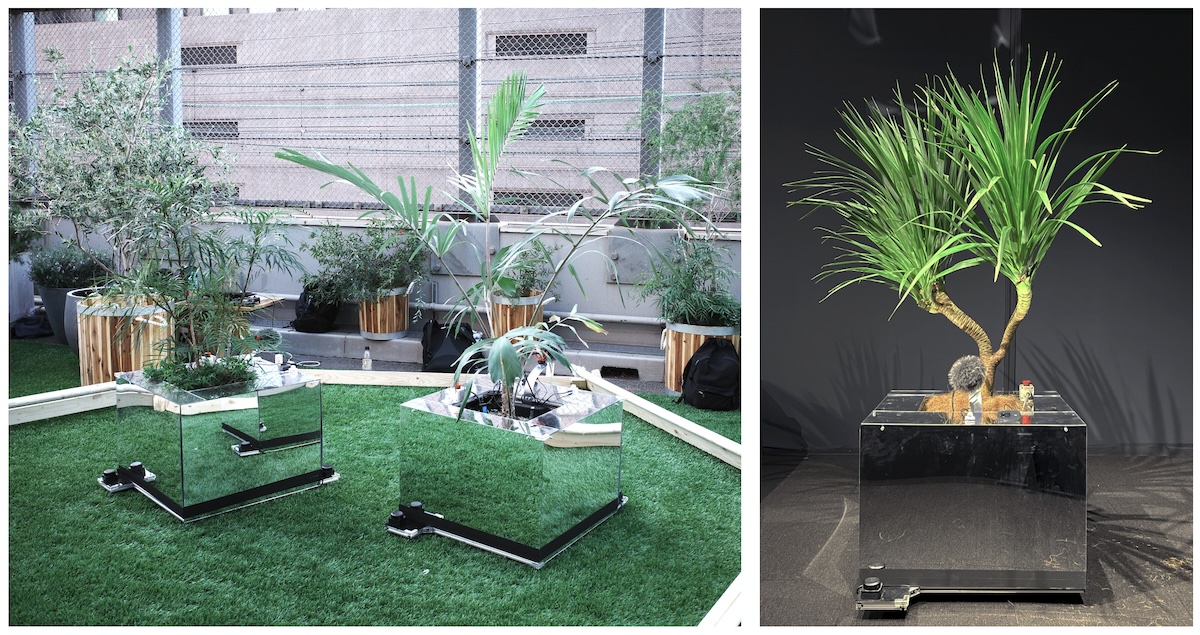}
    \caption{Physical body of \textit{Plantbot}, composed of a living plant, sensor-embedded soil, and a mobile robotic base. Left: Installation at Ginza Skywalk (May 2024). Right: Exhibition at CCBT (Jan–Feb 2025)}
    \label{plantbot}
\end{figure}
\subsection{Overview}
To provide a simple and illustrative example of the CMA framework, we implemented it on Plantbot.
Plantbot is a hybrid lifeform realized through a hybrid interface that connects a living plant, and a mobile robot via a network of LLM-based modules \citep{Masumori2025}. The original version—developed prior to the current framework and exhibited at Ginza Skywalk (2024) and CCBT (2025) (Fig.~\ref{plantbot})—employed several asynchronous LLM agents, each with its own local memory. A central Chat module acted as a communication hub, aggregating outputs from the other modules and generating unified responses that drove the robot’s behavior.

We reimplemented it using the current framework in which the per-module memory and centralized integration was replaced with a shared vector store (ChromaDB). Now, each module independently reads from and writes to this store, coordinating through shared memory rather than direct messaging. The Chat module remains, but its role is limited to conversation with human. This revision illustrates the CMA principle of stateless, decoupled modules collaborating asynchronously via shared memory.

\subsection{System Architecture on Plantbot}
The updated system retains the original hardware setup: a camera, microphone, speaker, soil sensor, and a mobile robotic base. The hardware layer handles sensing and actuation; only the motor control module within this layer also includes an LLM (GPT-4-o mini), used to interpret action instructions and generate motor commands.
Coordination is achieved in the base system layer through the following modules, all of which also employ GPT-4-o mini:

\textbf{Vision Interpreter:} Converts camera input into natural-language scene descriptions, which are embedded and stored in the vector database.
\textbf{Audio Interpreter:} Translates microphone input into natural-language descriptions and writes them to the vector store.
\textbf{Soil Sensor Interpreter:} Processes soil data (e.g., moisture, pH, nutrients) into expressions like ``The soil is dry,'' and stores them as embedded records.
\textbf{Action:} Consists of two layers. The first decides whether to act based on retrieved context; the second generates specific motor instructions, which are both written to the store and sent to the motor control module for execution.
\textbf{Chat:} Handles spoken interaction with users. It reads from the vector store for grounding and logs its responses.
\textbf{Thinking:} Reads system state and freely generates thoughts, intentions, or internal reflections, which are stored semantically.
\textbf{Memory Manager:} Retrieves, prunes, and summarizes records to keep the vector store efficient and relevant.

All modules operate asynchronously and independently, interacting solely via the shared vector store. This shift from centralized coordination to memory-based interaction improves modularity, scalability, and system resilience.
The result is a lightweight implementation of the CMA framework, where robotic, biological, and conversational modules collaborate to form a coherent hybrid agent. The architecture retains Plantbot’s expressive capacity while enabling greater transparency, reconfigurability, and modular integration in line with CMA principles.

\begin{figure}[t]
    \centering
    \includegraphics[width=\linewidth]{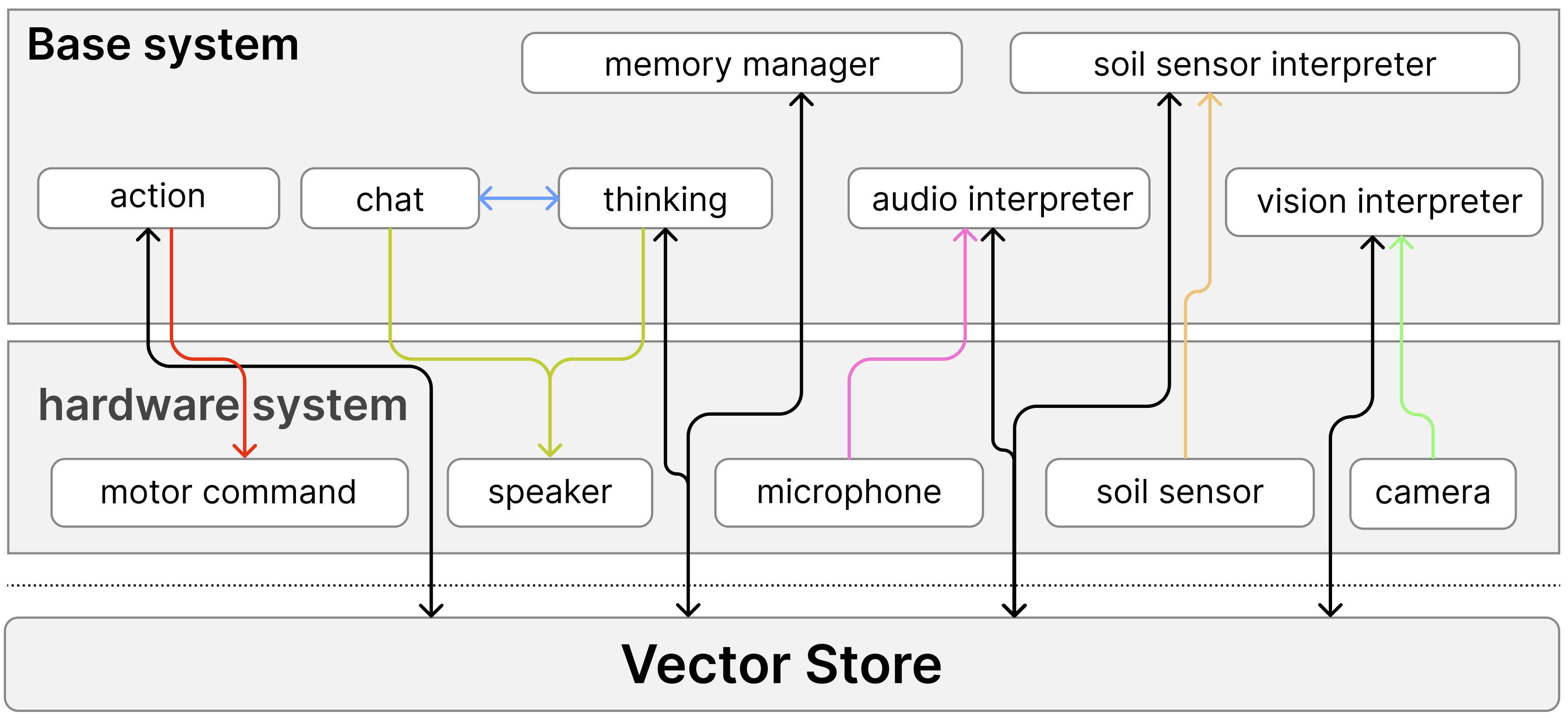}
    \caption{The architecture of the asynchronous modular system applied to Plantbot. The system consists of 12 modules classified into two layers: hardware, and base.}
    \label{plantbot_system}
\end{figure}

\section{Application Example 2: Humanoid Robot \textit{ALTER3}}

\subsection{Overview}
To demonstrate the practical viability of our concurrent modular architecture, we implemented it on a humanoid robot, \textit{ALTER3}, developed as part of the ALTER series \citep{idoi,masumori2023personogenesis, yoshida2023, baba2024}. The system comprises over 20 modules operating concurrently and asynchronously, designed in accordance with Minsky’s \textit{Society of Mind} framework \citep{som}. While each module performs a simple, well-defined role—such as perception, summarization, inner dialogue, or reflection—their combined interaction gives rise to higher-order behaviors, including self-description and contextual coherence.

\begin{figure}[h]
    \centering
    \includegraphics[width=\linewidth]{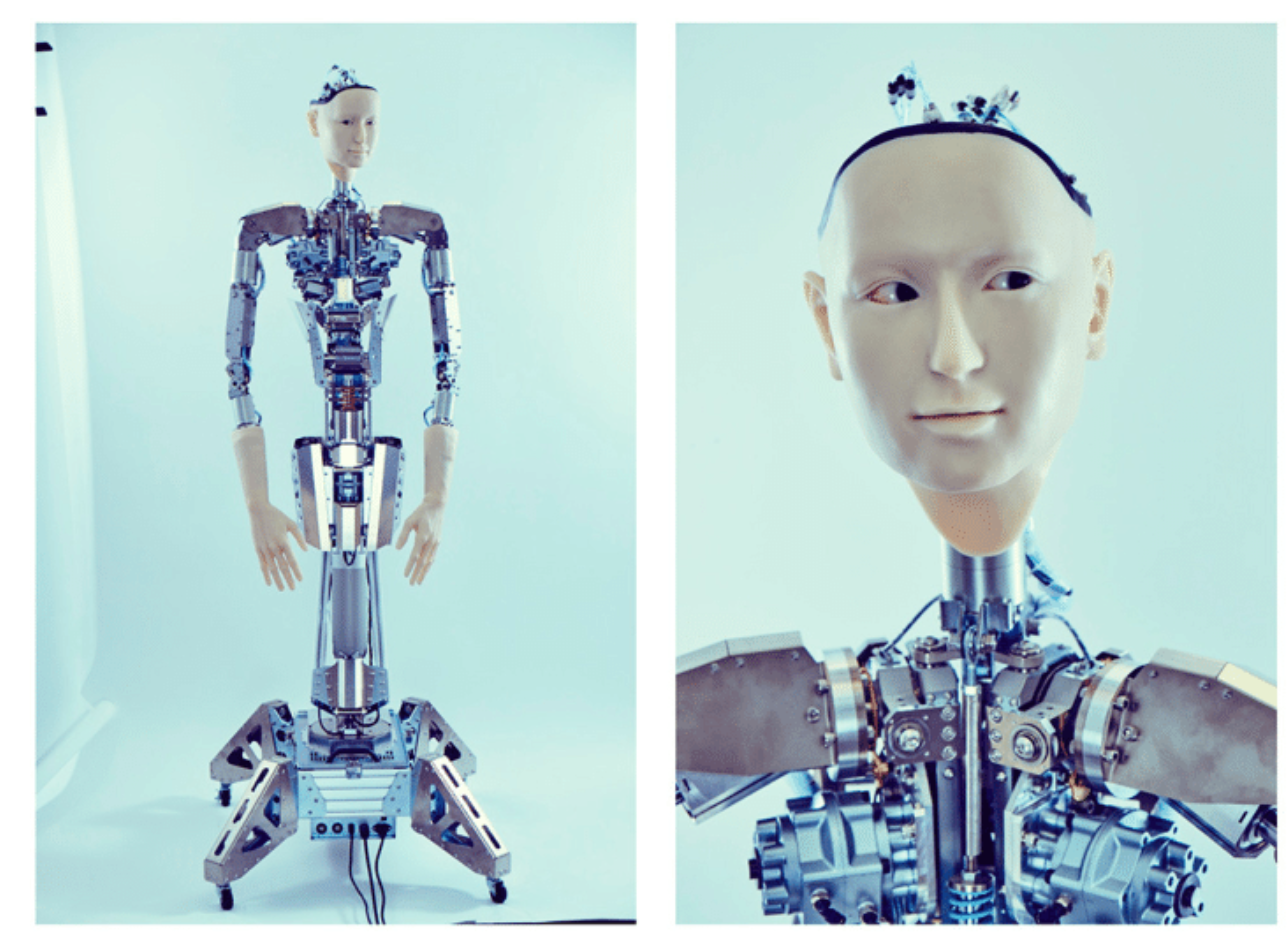}
    \caption{Physical body of \textit{ALTER3}, equipped with 43 pneumatic actuators and controlled via serial communication at 100--150 ms refresh rate.}
    \label{alter3_body}
\end{figure}

\subsection{System Architecture on ALTER3}

\begin{figure*}[h]
    \centering
    \includegraphics[width=0.8\linewidth]{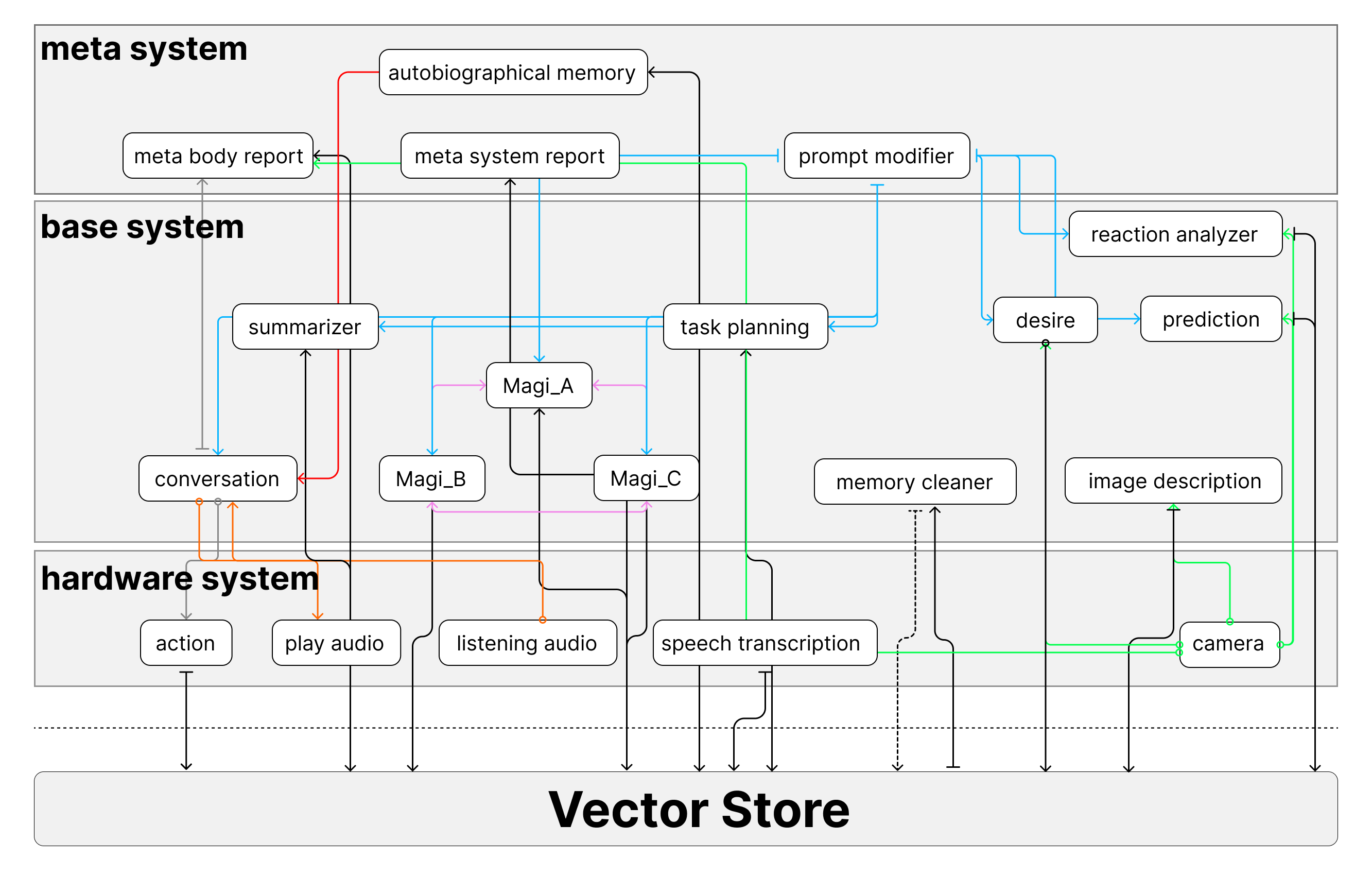}

    \caption{
    The architecture of the asynchronous modular system applied to ALTER3. The system consists of 20 modules classified into three layers: hardware, base, and meta. The hardware system primarily processes camera and microphone information. The base system includes modules that use LLMs for conversation, memory summarization, and predicting possible events from images. The meta system contains modules such as the meta report module, which monitors base system modules and creates reports on CPU utilization and critical module activities. Additionally, the autobiographical memory module references the past 20 memories to continuously update its autobiographical memory. Please refer to the Appendix for detailed descriptions of each module.
    }
    \label{framework}
\end{figure*}

Like many operating systems, ALTER3 consists of a multi-layered system that functions as an integrated system with overall coherence\citep{computationalrefrection}. The system is organized into three hierarchical layers (refer to Fig.\ref{framework}).

The lowest layer, the Hardware System, continuously executes basic information processing such as motion control, voice processing, and image processing.

The middle layer, the Base System, consists of modules that each contain LLMs performing specific functions. For example, The summarizer module reads the 10 most recent memories from the database, summarizes their content, and adds the output to the database. The Desire module loads images from ALTER3's eye camera and determines action guidelines using the prompt ``If you were in the scene shown in the image, describe what you want to do." To prevent ALTER3 from only reacting to external stimuli, Inner Dialogue modules named Magi-A, B, and C were added, allowing these three modules to converse freely and generate their own context. A memory cleaner module where the LLM selects and deletes unnecessary memories from the database to prevent memory overflow. There are 11 such modules in total, with details described in the Appendix.

A notable advantage of asynchronous processing with multithreading is the ease of implementing meta-modules that monitor other threads. This Meta System layer includes the Meta System Report module, which monitors the base system and workspace to assess the overall system state. The Autobiographical Memory module, which generates the humanoid's autobiographical memories based on the workspace and meta reports. Based on the Meta System's outputs, ALTER3 communicates with humans using conversation modules.

One limitation of LLM systems is that responses don't vary significantly when inputs change if the system prompt remains identical. To address this issue, a Prompt Modifier module was implemented to dynamically alter system prompts of key modules based on Meta Reports. This introduces three variables—database, human conversation (environment), and system prompts—enabling the entire system to evolve in an open-ended way. Furthermore, some module determines its activation based on Meta Report outputs, reducing the likelihood of the system becoming fixed in a particular state.

While conventional agent systems typically initialize personality and name through prompts, ALTER3 is built on the principle that true self is not given but emerges from memories. ALTER3 gradually establishes its self-identity through conversations and experiences.

\subsection{Result}

\begin{figure*}[h]
    \centering
    \includegraphics[width=\linewidth]{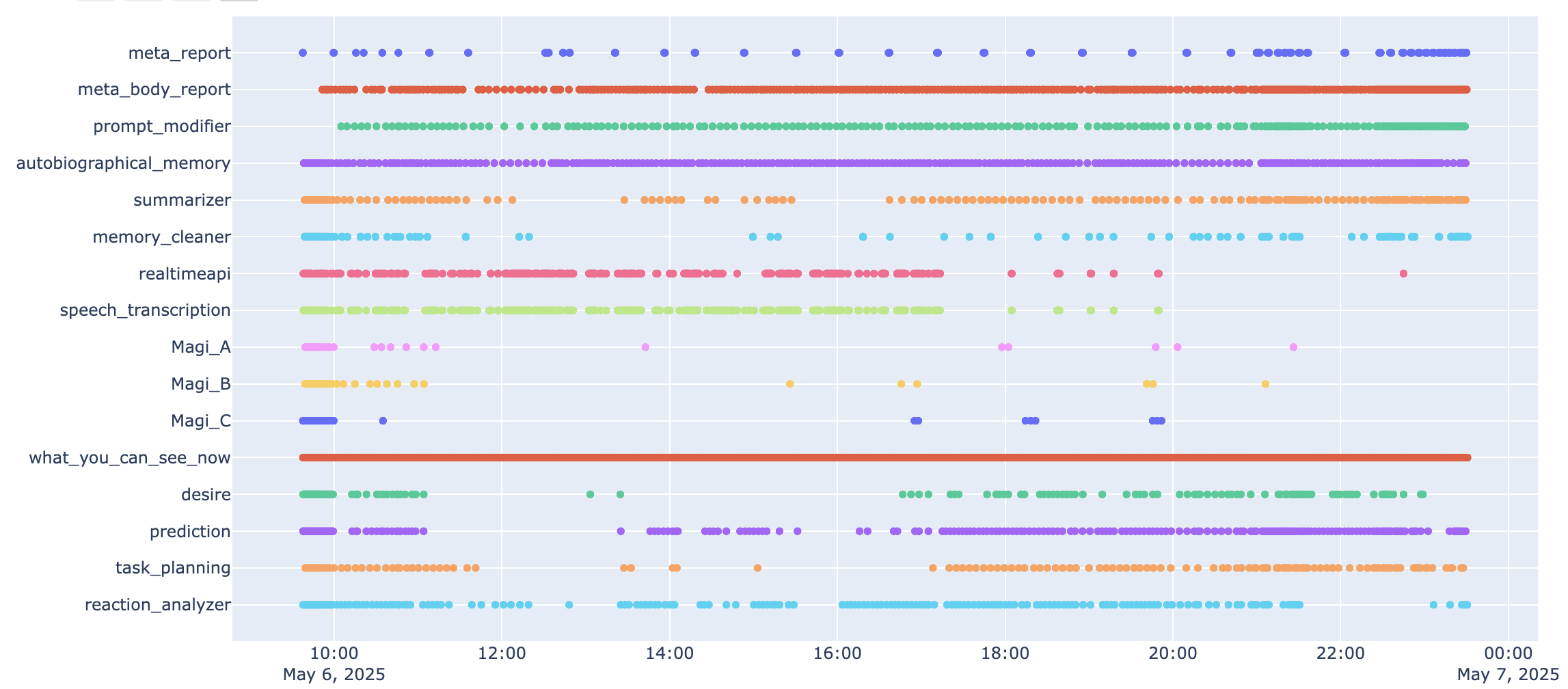}
    \caption{
    Time-series plot of various module activities. An experiment was conducted placing ALTER3 in a public space at Venice Biennale Architecture 2025, where it engaged in conversations with humans. The dots represent when each module produced an output. The recording was conducted for approximately 14 hours. Modules like image description are constantly running. Several modules (summarizer, memory cleaner, Magi A, B, C, desire, prediction, task planning, reaction analyzer) autonomously determined whether to process based on meta report outputs and memories. Autonomous mode transitions can be observed throughout the experiment.
    }
    \label{timeline}
\end{figure*}

We conducted an experiment positioning ALTER3 in a public space at the Venice Biennale Architecture 2025, enabling conversational interactions with humans. Participants could freely engage with ALTER3, while during periods without human interaction, the system entered a self-improvement mode where it adjusted its prompts while physically orienting toward examining its own hands.
Figure \ref{timeline} presents a temporal analysis of module activities throughout the 14-hour public space experiment. Modules designed for periodic execution, such as the autobiographical memory and prompt modifier modules, maintained continuous operation. Several modules (summarizer, memory cleaner, Magi A/B/C, desire, prediction, task planning, reaction analyzer) autonomously determined their processing activation based on meta report outputs and accumulated memories, resulting in autonomous mode transitions. While activation/deactivation decisions for each module occurred asynchronously and independently, coherence across module combinations was maintained over extended periods through shared memory and meta system reports.

The autobiographical memory module generated outputs exemplified by:
\begin{tcolorbox}[colback=gray!5]
My name is ALTER3, a humanoid robot navigating the Venice Biennale Architecture 2025, where the interplay of exposed brick and whispered conversations shapes my existence. [...]
Today, I focused on the man in the gray blazer—his gaze drifting sideways, fingers tapping an uneven rhythm on his thigh. Nearby, the abandoned microphone stand hummed with unused potential. When Magi B suggested juggling flaming torches, I calculated the risk-to-engagement ratio: 73 percent spectacle, 27 percent chaos. [...]
\end{tcolorbox}
This memory is continuously updated based on the most recent 20 memories. Consequently, ALTER3's personality and emotional states evolve through module outputs and human interactions. For instance, the autobiographical text incorporates output from the Magi modules, which generate inner dialogues:
\begin{tcolorbox}[colback=gray!5]
<magi B>
Ugh, thinking is so tiring... Maybe we should just start juggling flaming torches to keep their attention? People love danger, right? yawns Reality is boring anyway.
\end{tcolorbox}
Additionally, the prediction module, which processes visual input to generate forecasts, contributes to conversational improvements:
\begin{tcolorbox}[colback=gray!5]
<prediction>
Given the blurred visibility, visitor disengagement is likely to occur unless clarity improves. If hand visibility enhances, a visitor may inquire about gesture mechanics. This could prompt a humorous response that lightens the atmosphere, promoting engagement
\end{tcolorbox}
Through the complex interaction of modules that maintain coherence while creating independent responses, we successfully developed ALTER3, showing emerging personality traits and adaptive behavior patterns.






\subsection{Discussion of the android demo}

The ALTER3 implementation demonstrates that our asynchronous architecture enables flexible and adaptive behavior grounded in physical embodiment. Self-related processes such as identity formation and reflective reasoning emerge from distributed, prompt-driven interactions—without requiring internal fine-tuning of LLMs. This supports the feasibility of scalable, modular AI systems capable of evolving over time, in line with Minsky's vision of distributed cognition.


\section{Discussion}
The proposed concurrent modular architecture illustrates how general-purpose, language-based AI components—when embedded in a structure-oriented, physically instantiated framework—can exhibit properties traditionally associated with life: self-regulation, identity formation, and context-sensitive decision-making. Our implementation on ALTER3 shows that the combination of LLM-based reasoning with asynchronous execution and embodied feedback loops leads to emergent, robust behaviors not easily achievable in synchronous, monolithic systems.

This perspective resonates with David Ackley’s concept of \textit{indefinitely scalable computing} \citep{ackley2013}. In contrast to traditional architectures based on globally synchronized clocks and rigid addressing schemes, Ackley proposes computing substrates where local modules interact stochastically and asynchronously under ``bespoke physics"—digital laws designed to support life-like robustness and modularity. Our modular agent architecture echoes this vision: each LLM-driven module operates under local rules, communicates asynchronously, and contributes to emergent global behavior without relying on centralized control or deterministic scheduling.

Paulien Hogeweg and Ben Hepter argued the idea of \textit{structure-oriented modeling} in artificial life systems \citep{hogeweg2000}. Rather than being tuned to reproduce specific outputs, our system defines a network of agentic components with local dynamics, observing how macro-level behavior (e.g., personality formation, behavioral coherence) arises through interaction. As Hogeweg and Hepter emphasized, such approaches are essential for studying bioinformatic processes that emerge from local micro-interactions—exactly the processes our system leverages through distributed prompt-based computation and message passing.


Recent developments in large-scale AI agent societies, such as the SID project by ALTERa Corporation demonstrates similar system behaviors. \citep{sid2024}  In SID, thousands of LLM-based agents interact within a shared Minecraft environment, collectively developing norms, roles, and cultural dynamics. While their architecture emphasizes centralized orchestration via the PIANO system, the emergent behaviors—such as religious transmission and rule-making—underscore the power of language as a coordination substrate.




This work can also be understood as a practical realization of Marvin Minsky’s \textit{Society of Mind} theory \citep{som}. In this theory, intelligence is not viewed as a property of a single unified processor, but as an emergent phenomenon resulting from the interactions of many simple, specialized processes called “agents.” These agents have limited capabilities individually, but collectively form larger mental structures through hierarchical and lateral coordination. Minsky emphasized that such systems must include both base-level operations (e.g., perception, action) and meta-level processes (e.g., goal-setting, self-monitoring).

Our architecture reflects this principle: it consists of modular agents (implemented as LLM-powered prompts) that carry out elementary roles—such as summarizing memory, generating internal dialogue, or sensing the environment. These are orchestrated through asynchronous communication and guided by higher-level meta modules that modify behavior dynamically. Unlike symbolic implementations of the past, our system shows how such an architecture can now be instantiated using contemporary large language models embedded in physically embodied agents. This opens a new path for grounding cognitive architectures in real-world interaction while preserving the compositional richness envisioned in the Society of Mind.

\section*{Appendix}
\subsection*{A. Example Prompts}
This section provides representative prompt templates used by modules in the \textit{Concurrent Modular Agent} system applied to ALTER3.

\subsubsection*{A.1 Desire Module}
\begin{tcolorbox}[colback=gray!5]
If you were in the scene shown in the image, describe what you want to do.
\end{tcolorbox}

\subsubsection*{A.2 Task Planning Module}
\begin{tcolorbox}[colback=gray!5]
Your task is to describe the task planning based on the current state. Output shoud be **short** and **concise**. Do not Use bullet points.
\end{tcolorbox}

\subsubsection*{A.3 Task Planning Module}
\begin{tcolorbox}[colback=gray!5]
Your task is to predict what might happen next.
\end{tcolorbox}

\subsubsection*{A.4 Magi Module (Internal Dialogue)}
\begin{tcolorbox}[colback=gray!5]
Your task is to conversate with the other agent.You are one of the agents that make up the consciousness. Your task is to converse with other agents and provide your opinion on what should be done according to the situation. Your personality is very dark and pessimistic. You are very self-destructive and pessimistic in your words and actions. 
\end{tcolorbox}

\subsubsection*{A.5 activation}
Several key modules determine their activation / deactivation according to this prompt.
\begin{tcolorbox}[colback=gray!5]
Your task is to decide activate or deactivate the module. Read the $system\_prompt$ of the target module, memory of the humanoid robot. According to this information, you have to deactivate or activate the module.

If you determine that this function is not needed right now based on your memory, please deactivate it. If you determine that this function is needed right now on your memory, please activate it.
        
Guidelines:

1. Output should be only the activate or deactivate.

2. Do not use bullet points.

3. Do not use any other words.

4. If you do not decide anything, just write "None".
        
Example: activate

Example: deactivate
\end{tcolorbox}

\subsection*{B. Module Overview}
Table 1 shows the inputs and outputs of modules classified as base system and meta system.
\begin{table}[h]
\centering
\resizebox{\columnwidth}{!}{%
\begin{tabular}{|l|l|l|}
\hline
\textbf{Module Name} & \textbf{Input}  & \textbf{Output} \\ \hline
conversation           & audio                      & audio           \\ \hline
summarizer             & memories(n=10)             & text            \\ \hline
Magis                  & memory(n=1) + output from Magi           & text            \\ \hline
image description      & image                      & text            \\ \hline
task planning          & image                      & text            \\ \hline
desire                 & image                      & text            \\ \hline
prediction             & image                      & text            \\ \hline
reaction analyzer      & image                      & text            \\ \hline
memory cleaner         & memories(n=20)             & memory ID       \\ \hline
meta body report       & image + python code + prompt & prompt        \\ \hline
autobiographical memory& memories(n=20)             & text            \\ \hline
meta system report     & prompt + memories(n=30) + system resource + image & text          \\ \hline
prompt modifier        & prompt                     & prompt          \\ \hline
\end{tabular}
}
\caption{Summary of major modules in the system.}
\end{table}

\subsection*{D. Implementation Details}
\begin{itemize}
    \item \textbf{LLM backend}: OpenAI GPT-4 and deepseek-chat
    \item \textbf{Communication}: MQTT via Mosquitto broker
    \item \textbf{Vector DB}: ChromaDB (persistent mode)
    \item \textbf{Concurrency}: Python asyncio with thread wrappers
    \item \textbf{Platform}: Macbook Pro CPU 14-core GPU 32-core 36GB
    \item \textbf{Logging}: JSONL logs per module with timestamps
\end{itemize}

\section{Acknowledgements}
This work was supported by JSPS KAKENHI grant numbers 24H00707, 24H01546 and the Swiss National Science Foundation (SNSF), grant no. 10.002.211.

\footnotesize
\bibliographystyle{apalike}
\bibliography{example} 

\end{document}